\documentclass[runningheads]{llncs}
\usepackage{graphicx}

\usepackage{tikz}
\usepackage{comment}
\usepackage{amsmath,amssymb} 
\usepackage{color}

\usepackage[accsupp]{axessibility}  

\usepackage{tikz}
\usepackage{comment}
\usepackage{amsmath}
\usepackage{amssymb}
\usepackage{color}
\usepackage{cite}
\usepackage{soul}

\usepackage{url}
\usepackage{blindtext}
\usepackage{hyperref}

\usepackage{url}
\usepackage{booktabs, multirow} 
\usepackage{soul}
\usepackage{changepage,threeparttable} 
\usepackage{float}
\usepackage{pifont}
\usepackage{xcolor}
\usepackage{makecell}

\usepackage{graphicx}
\usepackage{float}
\usepackage{subfig}

\usepackage{caption}
\begin{document}
\pagestyle{headings}
\mainmatter
\def\ECCVSubNumber{9}  

\title{SimpleDG: Simple Domain Generalization Baseline without Bells and Whistles} 

\titlerunning{SimpleDG}
%
\author{Zhi Lv \and Bo Lin \and Siyuan Liang \and Lihua Wang \and\\  Mochen Yu \and Yao Tang \and Jiajun Liang} 
\authorrunning{Zhi Lv, Bo Lin et al.}
\institute{MEGVII Technology\\
\email{\{lvzhi,linbo,liangsiyuan,wanglihua,\\yumochen,tangyao02,liangjiajun\}@megvii.com}}
\maketitle
\begin{abstract}

We present a simple domain generalization baseline, which wins second place in both the common context generalization track and the hybrid context generalization track respectively in NICO CHALLENGE 2022. We verify the founding in recent literature, domainbed, that ERM is a strong baseline compared to recent state-of-the-art domain generalization methods and propose SimpleDG which includes several simple yet effective designs that further boost generalization performance. Code is available at \href{https://github.com/megvii-research/SimpleDG}{https://github.com/megvii-research/SimpleDG}.

\keywords{Domain Generalization, Domainbed, NICO++}
\end{abstract}

\section{Introduction}

\begin{figure}[t]
  \centering
\includegraphics[width=1.0\textwidth]{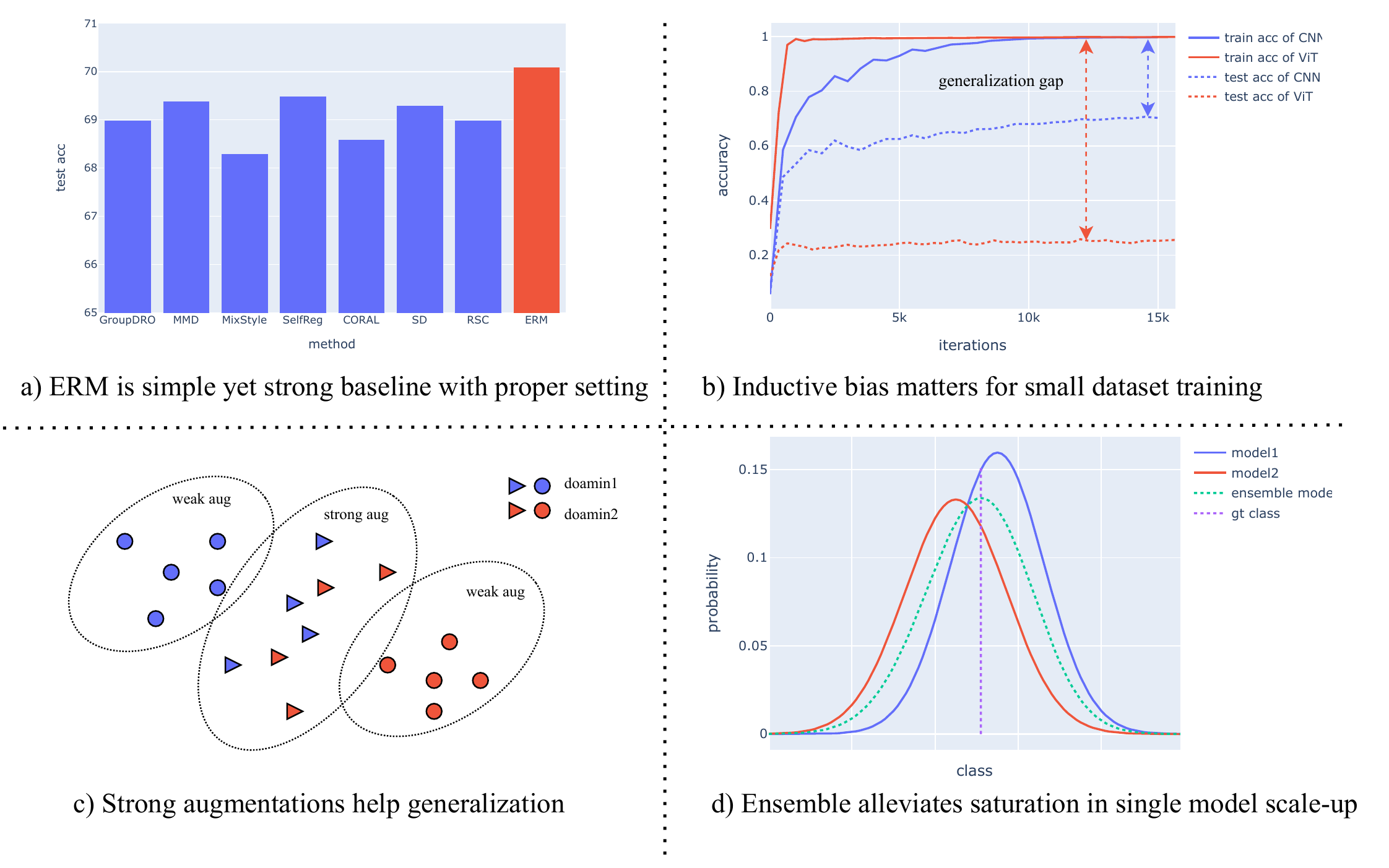}
  \caption{An overview of four key designs of our method. (a) ERM is a strong baseline when carefully implemented. Many modern DG methods fail to outperform it; (b) ViTs suffer from overfitting on the small dataset without pretraining. CNN due to its proper inductive bias has a much smaller train-test accuracy gap than ViTs; (c) Stronger augmentations help generalize better. The source domain's distribution is extended by strong augmentations and gets more overlap between different domains which is of benefit to optimizing for the target domain. (d) Models ensemble improves generalization performance as output probability distributions from models compensate for each other and results in more reasonable predictions.}
  \label{fig:intro_method}
\end{figure}

\subsection{Domain Generalization}
Deep learning models have achieved tremendous success in many vision and language tasks, and even beyond human performance in well-defined and constrained tasks. However, deep learning models often fail to generalize to out-of-distribution(OOD) data, which hinders greater usage and brings potential security issues in practice. For example, a self-driving car system could fail when encountering unseen signs, and a medical diagnosis system might misdiagnose with the new imaging system.

Aware of this problem, the research community has spent much effort in domain generalization(DG) where the source training data and the target test data are from different distributions. Datasets like PACS\cite{li2017deeper}, VLCS\cite{fang2013unbiased}, Office-Home\cite{venkateswara2017deep}, DomainNet\cite{peng2019moment} have been released to evaluate the generalization ability of the algorithms. Many methods like MMD\cite{li2018domain}, IRM\cite{arjovsky2019invariant}, MixStyle\cite{zhou2021domain}, SWAD\cite{cha2021swad} have been proposed to tackle the problem.

However, We find that traditional CNN architecture with simple technologies like augmentation and ensemble, when carefully implemented, is still a strong baseline for domain generalization problem. We call our method SimpleDG which is briefly introduced in Fiugre\ref{fig:intro_method}.

\subsection{DomainBed}
DomainBed\cite{gulrajani2020search} is a testbed for domain generalization including seven multi-domain datasets, nine baseline algorithms, and three model selection criteria.
The author suggests that a domain generalization algorithm should also be responsible for specifying a model selection method. Since the purpose of DG is to evaluate the generalization ability for unseen out-of-distribution data, the test domain data should also not be used in the model selection phase. Under this circumstance, the author found that Empirical Risk Minimization(ERM)\cite{vapnik1999overview} results on the above datasets are comparable with many state-of-the-art DG methods when carefully implemented with proper augmentations and hyper-parameter searching.


\begin{figure}[t]
  \centering
\includegraphics[width=1.0\textwidth]{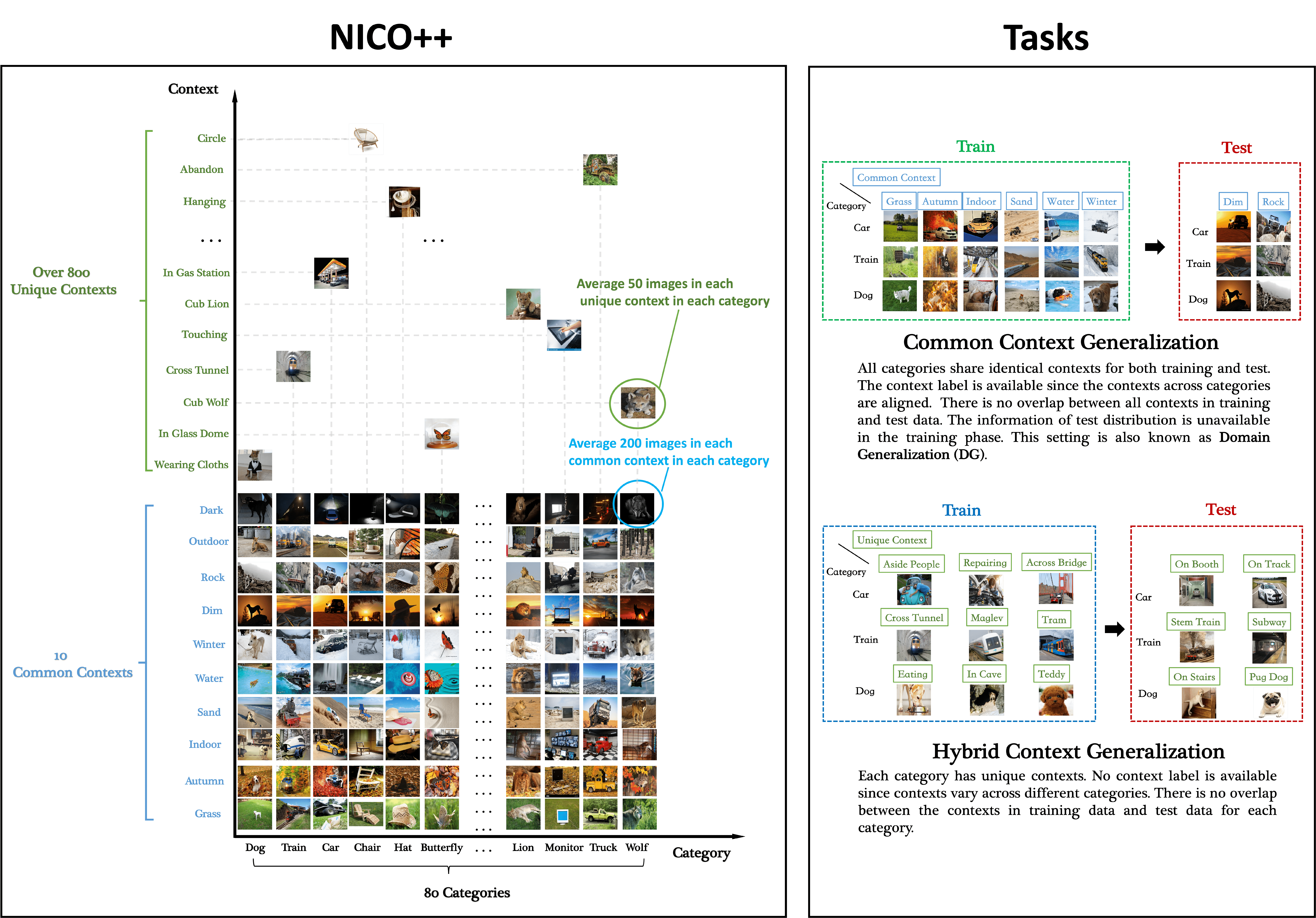}
  \caption{An overview of the NICO++ dataset. Identical contexts are shared across all different categories in the \textcolor{blue}{common contexts track}, while contexts are mutually exclusive in the \textcolor{green}{unique contexts track}. Some categories might have a single context only and therefore are more likely to suffer from overfitting problems.}
  \label{fig:intro_nico}
\end{figure}

\subsection{NICO++}
NICO++\cite{Zhang2022NICOTB} is a new DG dataset recently released in the NICO challenge 2022. The goal of the NICO Challenge is to promote research on discarding spurious correlations and finding the causality in vision. The advantages of the NICO++ dataset compared with popular DG datasets are as follows: 1) more realistic and natural context semantics. All the training data is collected from the real world and categorized carefully with specific concepts; 2) more diversity is captured in the dataset, which makes generalization not trivial; 3) more challenging settings. Except for the classic DG setting, NICO++ also includes  the unique contexts track where the overfitting problem is more severe.

The NICO Challenge contains two main tracks: 1) common context generalization track; 2) hybrid context generalization track. The difference between these two tracks is whether the context of all the categories is aligned and whether the domain label is available. Same as the classic DG setting, identical contexts are shared across all categories in both training and test data in the common context generalization track. However, contexts are mutually exclusive in the hybrid context generalization track as shown in Figure\ref{fig:intro_nico}. Context labels are available for the common context generalization track, but not for the hybrid context generalization track.

One main challenge of this competition comes from the unique context. Some of the samples have unique contexts, which might cause the model to overfit the unrelated background information. Another challenge comes from the small object and multi-object samples. As we observe, some samples in the dataset contain extremely small target objects.  It's very likely to crop the background part when generating the training data which may cause training noise and introduce bias. There are also some samples that have more than one target category. This may cause the model to be confused and overfit noise. The rule of preventing using extra data also makes the task harder since large-scale pre-trained models are not permitted.

\section{SimpleDG}
In this section, we  first introduce the evaluation metric of our experiments and then discuss four major design choices of our method named SimpleDG, including 
\begin{itemize}
  \item Why ERM is chosen as a baseline over other methods
  \item Why CNN is favored over ViT in this challenge
  \item How does augmentation help in generalization
  \item How to scale up the models to further improve performance
\end{itemize}

\subsection{Evaluation Metric}
To evaluate the OOD generalization performance internally, we use 4 domains, i.e. dim, grass, rock, and water, as the training domains(in-distribution) and 2 domains, i.e. autumn and outdoor, as the test domains(out-of-distribution). For model selection, we split 20\% of the training data of each domain as the validation dataset and select the model with the highest validation top-1 accuracy. All numbers are reported using top-1 accuracy on unseen test domains.\\

For submission, we retrain the models using all domains with a lower validation percentage(5\%) for both track1 and track2. Because we find that the more data we use, the higher accuracy we got in the public test dataset.

\subsection{Key Designs of SimpleDG}
\subsubsection{I. ERM is a simple yet strong baseline}
\hfill \break
A recent literature\cite{gulrajani2020search} argues that many DG methods fail to outperform simple ERM when carefully implemented on datasets like PACS and Office-Home, and proposes a code base, called domainbed, including proper augmentations, hyperparameter searching and relatively fair model selection strategies. 

We conduct experiments on NICO dataset with this code base and extend the algorithms and augmentations in domainbed. Equipped with our augmentations, we compare ERM with recent state-of-the-art DG algorithms. We find the same conclusion that most of them have no clear advantage over ERM as shown in Table\ref{SOTA_DG_COMPARE}.

\begin{table}[!htp]\centering
\caption{Many DG methods fail to outperform simple ERM}
\label{SOTA_DG_COMPARE}
\begin{tabular}{cc}
\hline
algorithm & test acc     \\ \hline
GroupDRO\cite{sagawa2019distributionally}  & 69.0 \\
MMD\cite{li2018domain}       & 69.4 \\
MixStyle\cite{zhou2021domain}  & 68.3 \\
SelfReg\cite{kim2021selfreg}   & 69.5 \\
CORAL\cite{sun2016deep}     & 68.6 \\
SD\cite{pezeshki2021gradient}	      & 69.3 \\
RSC\cite{huang2020self}       & 69.0 \\ \hline
ERM \cite{vapnik1999overview}      & \textbf{70.1}  \\ \hline
\end{tabular}
\end{table}

\subsubsection{II. ViTs suffer from overfitting on small training sets without pretraining}
\hfill \break
ViT\cite{dosovitskiy2020image} has shown growing popularity these years, and we first compare the performance of ViT with popular CNN in track1. We choose one CNN model, ResNet18, and two vision transformer model, ViT-B/32 and CLIP\cite{radford2021learning}. CNN outperforms ViT significantly when trained from scratch with no pre-trained weights. ViT achieves higher training accuracy but fails to generalize well on unseen test domains. We tried ViT training tricks such as LayerScale\cite{touvron2021going} and stochastic depth\cite{huang2016deep}. The test accuracy improves, but there is still a huge gap compared with CNN as shown in Table\ref{CNN_ViT}. On the contrary, the ViTs outperform CNN when using pre-trained weights and finetuning on NICO dataset. 

We surmise that ViTs need more amount of training to generalize than CNNs as no strong inductive biases are included. So we decide not to use them since one of the NICO challenge rules is that no external data (including ImageNet) can be used and the model should be trained from scratch. 

\begin{table}[!htp]\centering
\caption{Test domain accuracy of CNN and ViTs on NICO track1}\label{CNN_ViT}
\begin{tabular}{@{}rccc@{}}
\toprule
\multicolumn{1}{l}{} & ResNet18 & ViT-B/32 & CLIP \\ \midrule
w/ pretrain          & 81       & 87       & 90   \\
w/o pretrain         & 64       & 30       &      \\ \bottomrule
\end{tabular}
\end{table}

\subsubsection{III. More and stronger augmentation help generalize better}
\hfill \break

\begin{figure}[t]
  \centering
\includegraphics[width=1.0\textwidth]{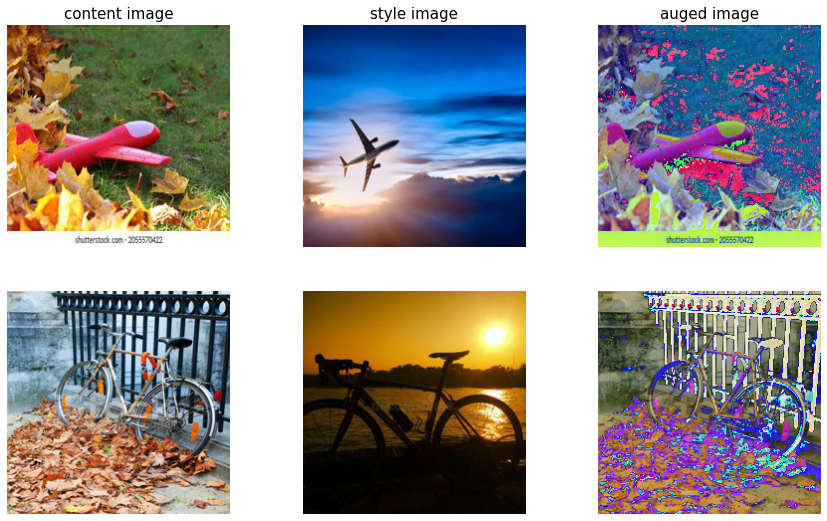}
  \caption{Visualization of Fourier Domain Adaptation. The low-frequency spectrum of the content image and the style image is swapped to generate a new-style image.}
  \label{fig:fda_viz}
\end{figure}

Both track1 and track2 suffer from overfitting since large train-validation accuracy gaps are clearly observed. Track2 has mutually exclusive contexts across categories and therefore suffers more from overfitting. With relatively weak augmentations, the training and test accuracy saturate quickly due to the overfitting problem. Generalization performance improves by adding more and stronger augmentations and applying them with a higher probability.

Following the standard ImageNet training method, we crop a random portion of the image and resize it to 224x224. We adopt timm~\cite{rw2019timm}'s RandAugment which sequentially applies N(default value 2) operations randomly chosen from a candidate operation list, including auto-contrast, rotate, shear, etc, with magnitude M(default value 10) to generate various augmented images. Test domain accuracy gets higher when more candidate operations(color jittering, grayscale and gaussian blur, etc.) are applied, and larger M and N are used.

Mixup~\cite{zhang2017mixup} and FMix~\cite{harris2020fmix} are simple and effective augmentations to improve generalization. Default Mixup typically samples mixing ratio $\lambda \approx \text{0 or 1}$, making one of the two interpolating images dominate the interpolated one. RegMixup\cite{pinto2022regmixup} proposes a simple method to improve generalization by enforcing mixing ratio distribution concentrate more around $\lambda \approx 0.5$, by simply increase the hyper-parameter $\alpha$ in mixing ratio distribution $\lambda \sim Beta(\alpha, \alpha)$. We apply RegMix to both Mixup and Fmix to generate augmented images with more variety. With these stronger augmentations, we mitigate the saturation problem and benefit from a longer training schedule.

\begin{table}[!htp]\centering
\caption{The breakdown effect for augmentation, high resolution finetuning and ensemble inference for Top-1 accuracy (\%) of NICO challenge track 1 training on ResNet-101.}\label{tab: 4} 
\scriptsize
\begin{tabular}{lccccc}\toprule
Method & Top-1 accuracy (\%) \\
Vanilla & 81.22 \\
+ RandAugment & 82.85  \\
+ Large alpha Mixup series & 84.58 \\
+ Fourier Domain Adaptation & 85.61 \\
+ High Resolution Fine-tune & 86.01 \\
+ Ensemble inference & \textbf{87.86} \\
\bottomrule
\end{tabular}
\end{table}

For domain adaption augmentation, we adopt Fourier Domain Adaptation~\cite{yang2020fda} proposed by Yang et al. FDA uses Fourier transform to do analogous ``style transfer". FDA requires two input images, reference and target images, it can generate the image with the ``style" of the reference image while keeping the semantic ``content" of the target image as shown in Figure\ref{fig:fda_viz}. The breakdown effect for each augmentation is shown in Table\ref{tab: 4}.\\

\subsubsection{IV. Over-parameterized models saturate quickly, and ensemble models continue to help}
\hfill \break

Big models are continuously refreshing the best results on many vision and language tasks. We investigate the influence of model capacity on NICO with the ResNet~\cite{he2016deep} series. When we test ResNet18, ResNet50, and ResNet101, the accuracy improves as the model size increases. But when we continue to increase the model size as large as ResNet152, the performance gain seems to be saturated. The capacity of a single model might not be the major bottleneck for improving generalization when only the small-scale dataset is available.

\begin{figure}[t]
  \centering
\includegraphics[width=1.0\textwidth]{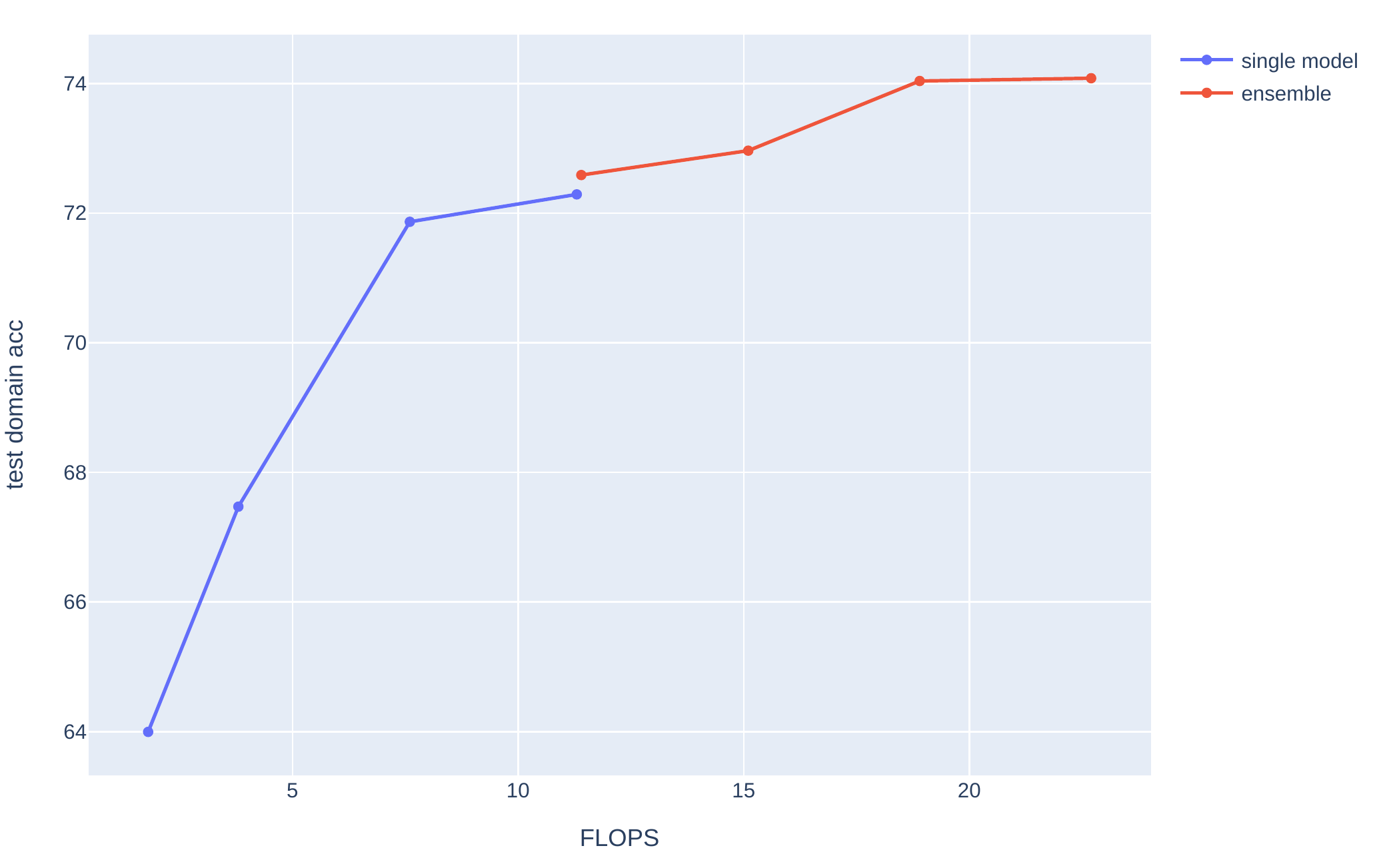}
  \caption{Test domain accuracy with different model size}
  \label{fig:exp_model_size}
\end{figure}

To further scale up the model, we adopt the ensemble method which averages the outputs of different models. When we average ResNet50 and ResNet101 as an ensemble model whose total flops is close to ResNet152, the performance gets higher than ResNet152. When further averaging different combinations of ResNet50, ResNet101, and ResNet152, the test accuracy get up to 2\% improvement. The ensemble method results are shown in Figure\ref{fig:exp_model_size}.

To figure out how ensemble helps, we conduct the following experiments. We first study ensemble models of best epochs from different train runs with the same backbone such as ResNet101. There is nearly no performance improvement even with a large ensemble number. The variety of candidate models should be essential for the ensemble method to improve performance. We launch experiments with different settings including different augmentations and different random seeds which influence the training/validation data split while still keeping the backbone architecture the same, i.e. ResNet101, among all experiments. This time, the ensemble models of these ResNet101s get higher test accuracy. We conclude that model variety not only comes from backbone architecture but also can be influenced by experiment settings that might lead to significantly different local minimums.

\subsection{More Implementation Detail}

\noindent \textbf{Distributed training.}\label{implementation} We re-implemented the training config using PyTorch's Distributed Data Parallels framework~\cite{li2020pytorch}. We can train ResNet101 with 512 batch-size in 10 hours with 8 GPUs(2080ti).

\noindent \textbf{Training from scratch.}\label{train_from_scratch} All models use MSRA initialization\cite{he2015delving}. We set the gamma of all batch normalization layers that sit at the end of a residual block to zero, SGD optimizer with 0.9 momentum, linear scaling learning rate, 1e-5 weight decay. We use 224x224 resized input images, 512 batch size, learning rate warmup for 10 epochs, and cosine annealing in overall 300 epochs. All experiments are trained with 8 GPUs.

\noindent \textbf{Fine-tune in high-resolution.}\label{finetune} We fine-tune all models in 448x448 resolution and 128 batch size for 100 epochs, this can further boost the model performance.

\noindent \textbf{Ensemble inference.}\label{ensemble_inference} In the inference phase, we use the ensemble method to reduce the variance. We average the features before softmax layer from multiple models, which is better than logits averaging after softmax layer.

\subsection{Public Results}
The top-10 public test dataset, which is available during the competition, results in track1 and track2 are shown in the Table \ref{public_result}.

\begin{table}[!htp]\centering
\caption{Top-10 Teams' Public Dataset Results of Track1 and Track2}\label{public_result}
\scriptsize
\begin{tabular}{lrrrrr}\toprule
\multirow{2}{*}{Rank} &\multicolumn{2}{c}{Track1} &\multicolumn{2}{c}{Track2} \\\cmidrule{2-5}
&Team &Top1-Acc &Team &Top1-Acc \\\midrule
1 &detectors\_218 &88.15704 &PingPingPang &84.62421 \\
2 &megvii\_is\_fp\textbf{(Ours)} &87.85504 &vtddggg &84.05049 \\
3 &mcislab840 &87.53865 &timmy11hu &81.99656 \\
4 &ShitongShao &86.83397 &megvii\_is\_fp\textbf{(Ours)} &81.49168 \\
5 &MiaoMiao &85.83447 &Wentian &79.91968 \\
6 &Wentian &85.75538 &czyczyyzc &79.35743 \\
7 &test404 &85.54685 &wangyuqing &78.81813 \\
8 &peisen &85.46775 &Jarvis-Tencent-KAUST &78.78371 \\
9 &HuanranChen &84.92126 &Wild &78.41652 \\
10 &wangyuqing &84.6624 &peisen &77.80838 \\
\bottomrule
\end{tabular}
\end{table}

\subsection{Private Results}
The NICO official reproduced our method and tested it on the private test dataset, which is unavailable during the competition, and the results are shown in Table\ref{private_result}.

Our method is quite stable between public dataset and private dataset, the ranking stays the same in track1 and becomes better in track2 while other methods undergo ranking turnover.

\begin{table}[!htp]\centering
\caption{Top-5 Teams' Private Dataset Results of Track1 and Track2}\label{private_result}
\scriptsize
\begin{tabular}{llcccr}\toprule
&Team &Phase 1 Rank &Phase 2 Score &Phase 2 Rank \\\midrule
\multirow{5}{*}{Track1} &MCPRL-TeamSpirit &1 &0.7565 &1 \\
&megvii-biometrics\textbf{(Ours)} &2 &0.7468 &2 \\
&DCD404 &6 &0.7407 &3 \\
&mcislab840 &3 &0.7392 &4 \\
&MiaoMiao &4 &0.7166 &5 \\ \hline
\multirow{5}{*}{Track2} &vtddggg &2 &0.8123 &1 \\
&megvii-biometrics\textbf{(Ours)} &4 &0.788 &2 \\
&PingPingPangPangBangBangBang &1 &0.7631 &3 \\
&jarvis-Tencent-KAUST &5 &0.7442 &4 \\
&PoER &8 &0.6724 &5 \\
\bottomrule
\end{tabular}
\end{table}


\section{Conclusion}


In this report, we proposed SimpleDG which wins both the second place in the common context generalization track and the hybrid context generalization track of NICO CHALLENGE 2022. With proper augmentations and a longer training scheduler, the ERM baseline could generalize well on unseen domains. Many existing DG methods failed to continue to increase the generalization from this baseline. Based on ERM, both augmentations and model ensembles played an important role in further improving generalization. 

After participating in the NICO challenge, we found that simple techniques such as augmentation and model ensemble are still the most effective ways to improve generalization. General and effective domain generalization methods are in demand, but there is still a long way to go.

\par\vfill\par
\clearpage
%
%
\bibliographystyle{unsrt}
\bibliography{egbib}
\end{document}


\definecolor{commentcolor}{RGB}{3, 148, 252}   
\newcommand{\PyComment}[1]{\ttfamily\textcolor{commentcolor}{\# #1}}  
\newcommand{\PyCode}[1]{\ttfamily\textcolor{black}{#1}} 
\pagestyle{headings}
\mainmatter
\def\ECCVSubNumber{807}  

\title{Supplementary Materials of \\Efficient One Pass Self-distillation with Zipf's Label Smoothing} 

\titlerunning{Zipf's LS}
%

\titlerunning{Zipf's LS}
%
\author{Jiajun Liang\orcidlink{https://orcid.org/0000-0001-5586-340X} \and Linze Li\orcidlink{https://orcid.org/0000-0001-9091-8699} \and Zhaodong Bing \and Borui Zhao \and Yao Tang \and Bo Lin \and Haoqiang Fan}
\authorrunning{Jiajun Liang\orcidlink{https://orcid.org/0000-0001-5586-340X} et al.}
\institute{MEGVII Technology
\email{\{liangjiajun,lilinze,bingzhaodong,zhaoborui,tangyao02,linbo,fhq\}@megvii.com}}
\maketitle

\section{Explanation to empirical observation}

We find that the Zipf's prior could help
generate non-uniform supervision for non-target classes
in a one-pass way. In this section, we provide a simple intuition to explain why Zipf's law should occur for predictions from multi-class classification.

We postulate that one main source of the non-zero network predictions is the inevitable non-orthogonality of the inter-class feature vectors as more and more classes are packed into the finite-dimensional feature space. In a simplified model, we assume that the decision vectors corresponding to each class are uniformly distributed on a high-dimensional unit sphere. Then for another random query vector on the sphere, their inner-products with it distribute in the shape of a Gaussian when the dimension is high enough. 
\begin{figure}[btp]
  \centering
\includegraphics[width=0.4\textwidth]{figs/ideal.png}
  \caption{softmax Gaussian logits fit Zipf's law well}
  \label{fig:ideal fit}
\end{figure}

We propose a toy experiment to verify that softmax Gaussian logits fit Zipf's law well. As shown in Algorithm \ref{gaussian_logits}, first, we sampled random vectors from multivariate normal distribution $\mathcal{N}(\boldsymbol{0},\,I_{1000})$ as logits of different samples. Then logits for each sample are sorted and applied with softmax to get probabilities. At last, we average the sorted probabilities across all samples and plot the probabilities-rank relation for the top 32 categories in log-log space. It could be seen that a straight line pattern shows up in Figure \ref{fig:ideal fit}.



\begin{algorithm}[h]
\SetAlgoLined
    \PyComment{generate Gaussian logits,1000 samples, 1000 classes} \\
    \PyCode{logits = np.random.randn(1000, 1000)} \\
    \PyComment{sort in class dimension} \\
    \PyCode{sorted\_logits = np.sort(logits, axis=1)} \\
    \PyComment{probability predictions by applying soft-max on the logits} \\
    \PyCode{sorted\_preds = np.exp(sorted\_logits) / np.sum(np.exp(sorted\_logits),axis=1)[:,None]}\\ 
    \PyComment{averaged across samples} \\
    \PyCode{mean\_sorted\_preds = np.mean(sorted\_preds, axis=0)} \\
    \PyComment{top 32 ranks considered} \\
    \PyCode{top32\_sorted\_preds = mean\_sorted\_preds[::-1][:32]} \\

\caption{simulation of ranking softmax Gaussian logits}
\label{gaussian_logits}
\end{algorithm}

We also compare several most frequently-used distributions of long-tail shapes (Zipf's law, exponential and log-normal) to fit the empirical softmax scores, as shown in Table \ref{tab:laws_of_fit}. All parameters of distributions are estimated by most-likelihood estimation. Common statistical test metrics such as R square are measured. Zipf's law outperforms the other two in all kinds of metrics.

\setlength{\tabcolsep}{6pt}
\begin{table}[!htp]\centering
\caption{Statistical test of how well different distributions fit on empirical averaged predictions on INAT-21. The top 50 categories are considered. For tests such as Chisquare and Kolmogorov–Smirnov which heavily rely on the amount of the samples, we sample $10^5$ instances from the empirical distribution. $D$ is the Kolmogorov–Smirnov statistic and $p$ is the p-value. Zipf's law outperforms the other two by all kinds of metrics.}\label{tab:laws_of_fit}
\scriptsize
\begin{tabular}{lcccc}\toprule
\textbf{Metric} &\textbf{Zipf's law} &\textbf{Exponential} &\textbf{Log-normal} \\\cmidrule{1-4}
$R^2$ &0.99992 &0.6768 &0.9672 \\\cmidrule{1-4}
Kullback–Leibler divergence &0.0000667 &0.315 &0.0219 \\\cmidrule{1-4}
Jensen–Shannon divergence &0.0000167 &0.063 &0.00544 \\\cmidrule{1-4}
Chisquare &13.3 &451677 &4499 \\\cmidrule{1-4}
Kolmogorov–Smirnov &$D$=0.00278 &$D$=0.265 &$D$=0.0823 \\
 & $p$=0.42 &$p$=0.0 &$p$=0.0 \\
\bottomrule
\end{tabular}
\end{table}



\setlength{\tabcolsep}{6pt}
\begin{table}[ht]\centering
\caption{The detail of hyper parameters for different datasets.}\label{tab: hyper_params}
\scriptsize
\begin{tabular}{lccccc}\toprule
Dataset &$\lambda$ &$\alpha$ &dense layer &$\beta$ \\\cmidrule{1-5}
CIFAR100 &0.1 &1.0 &2 &0.1 \\
TinyImageNet &1.0 &1.0 &2 &0.5 \\
ImageNet &0.1 &1.0 &1 &/ \\
INAT21 &1.0 &1.0 &1 &/ \\
\bottomrule
\end{tabular}
\end{table}

\section{More Experiment Details and Discussion}

\subsection{Hyperparameters}
\textbf{Hyperparameters setting rules.} Table \ref{tab: hyper_params} shows the detail of hyperparameters settings for different tasks. $\alpha$ controls the decay shape of
Zipf’s distribution, and is set to 1.0 in all tasks. $\lambda$ controls the regularization strength, which is set to 0.1 for CIFAR100 and ImageNet, and 1.0 for TinyImageNet and NAT21. For datasets with large-resolution inputs, such as ImageNet and INAT, using the final dense feature maps would be sufficient, and no more dense layers are required. For low-resolution tasks such as CIFAR100 and TinyImageNet, we use one more dense layer to get enough votes for dense ranking. In this case, we need $\beta$ to weigh the cross-entropy loss for learning the extra classifier. $\beta$ is set to 0.1 and 0.5 respectively on CIFAR100 and TinyImageNet. \\

\noindent \textbf{Hyperparameters ablation study.} 1) $\alpha$ is not sensitive where $\alpha\in[0.5, 1.5]$. 2) $\lambda$ is to control regularization strength and is positively correlated with the train/val acc gap. For tasks that are prone to overfitting(TinyImageNet and INAT whose train/val gap are $35\%$ and $25\%$), $\lambda$ is $1.0$. For tasks that are less overfitting(ImageNet train/val gap is $4\%$), $\lambda$ is $0.1$. 3) $\beta$ is optional and only recommended for small resolution tasks. It should be less than $0.5$ to avoid shadow learning of deeper layers. See Table \ref{tab:params_ablation} for details.

\begin{table}[!htp]\centering
\caption{Ablation study of hyperparameters $\alpha$,$\lambda$ and $\beta$ }\label{tab:params_ablation}
\scriptsize
\begin{tabular}{ccccccc}\toprule
$\alpha$ & 0.1 & 0.5 & 1.0 & 1.5  & 2.0 &\\\midrule
CIFAR100 & 77.21±0.29 &77.26±0.13 & \textbf{77.38±0.32} & 77.12±0.24 & 76.45±0.12 & \\
TinyImageNet & 58.85±0.16 & 59.06±0.21 & \textbf{59.25±0.20} & 58.64±0.18 & 53.35±0.41 & \\
\midrule\midrule
$\lambda$ & 0.01 & 0.1 & 0.5 & 1.0  & 1.5 &\\\midrule
CIFAR100 & 76.59±0.15 & \textbf{77.38±0.32} & 76.62±0.24 & 76.79±0.04 & 76.92±0.17 & \\
TinyImageNet & 56.86±0.36 & 57.65±0.01 & 58.41±0.17 & \textbf{59.25±0.20} & 58.03±0.27 & \\
\midrule\midrule
$\beta$(optional) & 0.05 & 0.1 & 0.3 & 0.5  & 0.7 &\\\midrule
CIFAR100 & 77.24±0.22 & \textbf{77.38±0.32} & 76.75±0.31 & 76.39±0.16 & 76.49±0.24 & \\
TinyImageNet & 58.71±0.17 & 58.77±0.18 & 59.48±0.31 & \textbf{59.25±0.20} & 59.00±0.12 & \\
\bottomrule
\end{tabular}
\end{table}


\subsection{SNR of Ranking}
Ranking the classes accurately is a key factor to generate proper Zipf's law distribution for the sample. In the method section, we propose a finer ranking method named dense classification ranking which exploits spatial classification results from the last few feature maps. A consequence of
this voting-based method is that we have to clip the Zipf’s
values to a uniform one after a certain rank, as they would
not receive sufficient votes to be distinguished individually.
To illustrate that ranking only a few top non-target classes is sufficient, we study the signal-to-noise ratio of rankings. The signal and noise of specific rank $r$ are calculated as the average and standard deviation among $r$-th probabilities from different sorted samples respectively. As shown in Figure.\ref{fig:snr}, we plot the SNR-rank curve on the INAT21 dataset, only the top 40 out of 10000 ranks whose SNR is larger than one. It's a good trade-off to just give power-law decayed probabilities to the top-ranking class since the SNR of tailing ranks is too low to give reliable ranks. 

\begin{figure}[t]
    \centering
    \subfloat[SNR-Rank relation across all classes]{
    \includegraphics[width=0.4\textwidth]{figs/snrleft.jpg}
    }\hskip2em
    \subfloat[SNR-Rank relation across top 40 classes]{
    \includegraphics[width=0.4\textwidth]{figs/snrright.jpg}
    }
    \caption{SNR-Rank relation plot for trained ResNet-50 model in INAT21 dataset. It can be seen that only the top 40 rankings have SNR larger than 1, which makes it hard to give reliable ranks for tailing class on the fly.
    }
    \label{fig:snr}
\end{figure}



\subsection{More Zipf's Soft Label Examples}
Figure~\ref{fig:result_more} illustrates more results of the top-5 predictions of our proposed method compared with the baseline method. The top three rows are sampled from ImageNet while the bottom three rows are sampled from INAT21.
It can be seen that: 
(1) There are several categories similar to the target class shown up in Zipf's soft labels, which provide meaningful label representations for the network to better grasp the concept of the target class.
(2) Fine-grained categories of the target class emerge in Zipf's soft labels, which can provide similar ``dark knowledge'' as knowledge distillation. \\

\subsection{Generalization: Performance on downstream tasks}
To measure the power of generalization of Zipf's LS, we conducted the transfer learning task by fine-tuning ImageNet pre-trained models on MS-COCO, as shown in Table \ref{tab:COCO}. Besides, we visualize loss landscapes\cite{li2018visualizing} of several efficient teacher-free methods(see Fig \ref{fig:landscape}), Zipf's LS achieves more flat convergence, which is a possible hint for better generalization\cite{minima3}.

\begin{table}[!htp]\centering
\caption{ImageNet pretrained ResNet50 for object detection}\label{tab:COCO}
\scriptsize
\begin{tabular}{lccccc}\toprule
Method & Vanilla(CE) &TF-KD & PS-KD& Zipf's LS(Ours)  & \\\midrule
AP &36.4\% & 36.4\%  & 36.5\%  &\textbf{36.6\%}\\
AP@0.5 &58.3\%&  56.7\% & 56.7\%&\textbf{58.8\%} \\
\bottomrule
\end{tabular}
\end{table}

\begin{figure*}[t]
  \centering
\includegraphics[width=1.\textwidth]{figs/loss_landscape.png}
  \caption{Comparison of loss landscape with efficient teacher-free methods}
  \label{fig:landscape}
\end{figure*}

\begin{figure}[hbp]
  \centering
\includegraphics[width=1.0\textwidth]{figs/result_more.pdf}
  \caption{Top-5 predictions visualization of our proposed method (Zipf's label smoothing) compared with the baseline method (cross-entropy). 
  The dark green, light green and red denote ground-truth, similar and irrelevant categories respectively.
  ``GT'' denotes the ground truth label and thus the hard label of the baseline method. 
  The baseline prediction is acquired under the supervision of the hard label and misclassified on the samples. 
  Our method exploits target-relevant categories to better represent the image, and obtains better results.
  }
  \label{fig:result_more}
\end{figure}

\par\vfill\par
\clearpage
%
%
\bibliographystyle{splncs04}
\bibliography{egbib}